\documentclass[10pt,twocolumn,letterpaper]{article}

\usepackage{cvpr}
\usepackage{times}
\usepackage{epsfig}
\usepackage{graphicx}
\usepackage{amsmath}
\usepackage{amssymb}
\usepackage{booktabs}
\usepackage{multirow}
\usepackage[table,xcdraw]{xcolor}

\usepackage[pagebackref=true,breaklinks=true,letterpaper=true,colorlinks,bookmarks=false]{hyperref}

\begin{document}

\title{Robust Facial Reactions Generation: An Emotion-Aware Framework with Modality Compensation}

\author{Guanyu Hu \\
Xi'an Jiaotong University\\
Queen Mary University of London\\
\and
Jie Wei\\
Xi’an Jiaotong University\\
\and
Siyang Song\\
University of Leicester\\
\and
Dimitrios Kollias\\
Queen Mary University London\\
\and
Xinyu Yang \\
Xi'an Jiaotong University\\
\and
Zhonglin Sun\\
Queen Mary University of London\\
\and
Odysseus Kaloidas\\
London School of Economics\\
}

\maketitle
\thispagestyle{empty}

\begin{abstract}

The objective of the Multiple Appropriate Facial Reaction Generation (MAFRG) task is to produce contextually appropriate and diverse listener facial behavioural responses based on the multimodal behavioural data of the conversational partner (i.e., the speaker). Current methodologies typically assume continuous availability of speech and facial modality data, neglecting real-world scenarios where these data may be intermittently unavailable, which often results in model failures. Furthermore, despite utilising advanced deep learning models to extract information from the speaker's multimodal inputs, these models fail to adequately leverage the speaker's emotional context, which is vital for eliciting appropriate facial reactions from human listeners. To address these limitations, we propose an Emotion-aware Modality Compensatory (EMC) framework. This versatile solution can be seamlessly integrated into existing models, thereby preserving their advantages while significantly enhancing performance and robustness in scenarios with missing modalities. Our framework ensures resilience when faced with missing modality data through the Compensatory Modality Alignment (CMA) module. It also generates more appropriate emotion-aware reactions via the Emotion-aware Attention (EA) module, which incorporates the speaker's emotional information throughout the entire encoding and decoding process. Experimental results demonstrate that our framework improves the appropriateness metric FRCorr by an average of 57.2\% compared to the original model structure. In scenarios where speech modality data is missing, the performance of appropriate generation shows an improvement, and when facial data is missing, it only exhibits minimal degradation.

\end{abstract}

\vspace{-0.5cm} 
\section{Introduction}
In recent years, the field of human-computer interaction has increasingly focused on developing systems capable of comprehending and appropriately responding to human conversations. This progression is crucial for enabling more natural and effective communication between humans and machines. A pivotal aspect of this domain involves enabling virtual agents and robots to autonomously generate realistic, human-like facial reactions that are both contextually appropriate and diverse in response to their human interlocutors' multi-modal behaviours, typically encompassing speech and facial expressions \cite{whatwhyhow,vico,song2022learning,shao2021personality,ng2022learning,10193422}.

Recent developments in Multiple Appropriate Facial Reaction Generation (MAFRG) have overcome the limitations of traditional one-to-one mapping methods, which struggled with the variability in appropriate reactions. By reframing the problem as a mapping from speaker behaviour to a distribution of appropriate reactions, these models have enhanced the diversity and synchrony of generated facial reactions \cite{luo2023reactface,liang2023unifarn,nguyenvector,liuone,yu2023leveraging,nguyenmultiple}.

Despite significant advancements in leveraging multi-modal inputs, current methodologies often fall short in two critical areas. Firstly, existing models typically assume the constant availability of all modalities—audio and visual. This assumption does not hold in real-world scenarios where one or more modalities might be intermittently unavailable due to factors such as network issues, environmental noise, or occlusions \cite{ma2021smil} \cite{wang2023multi}. Secondly, there is insufficient emphasis on the speaker's emotional context, which is crucial for generating appropriate listener reactions \cite{emotionallistener} \cite{lin2019moel}. Effectively incorporating emotional cues into the facial reaction generation process is vital for the authenticity and relevance of the produced responses \cite{gan2023emotalkhead} \cite{xu2023high}. The failure to address these challenges results in systems that are less effective and robust in practical applications. There is a clear need for a framework that can both gracefully handle missing modality information and effectively integrate emotional cues.

In this paper, we propose Emotion-aware Modality Compensatory (EMC) framework designed to address the challenges of generating emotion-aware reactions that remain robust even when faced with incomplete modality data. To handle the challenge of missing modality data in real-world applications, we introduce a Compensatory Modality Alignment (CMA) Module. This module compensates for missing modality data by aligning and complementing the available data, enabling the system to infer and reconstruct missing information based on what is present. Additionally, we introduce an Emotion-aware Attention (EA) Module, which dynamically integrates the speaker's emotional information into their multimodal data. This ensures that the emotional state of the speaker is accurately understood and appropriately reflected in the generated reactions. By incorporating these modules, our framework ensures that the system can maintain high performance and generate appropriate responses even when some modalities are absent. This versatile solution is designed to be seamlessly integrated into existing models,  preserving their inherent advantages while significantly enhancing their generation appropriateness and robustness in scenarios with modality data missing. 

To validate our approach, we conduct extensive experiments and evaluations. The results demonstrate the effectiveness of our framework in producing contextually appropriate and emotionally aware reactions, even in the presence of incomplete modality information. In summary, the main contributions of this paper are summarised as follows:
\begin{itemize}
    \item \textbf{Emotion-aware Modality Compensatory (EMC) Framework}: We propose a novel framework that integrates speaker emotional cues throughout the entire encoding and decoding process, facilitating the virtual agent to better emotional understanding and generating contextually appropriate facial reactions, even with incomplete modality data.

    \item \textbf{Compensatory Modality Alignment (CMA) and Emotion-aware Attention (EA) Modules}: The CMA module aligns and compensates for missing modality information, enabling the model to maintain high performance and generate robust reactions. Additionally, the EA module dynamically incorporates the speaker's emotional context into various feature formats within the model's information flows, ensuring a comprehensive and nuanced understanding of the input data.

    \item \textbf{Extensive experiments and evaluations}: Results demonstrate the framework's effectiveness and robustness in producing emotionally aware reactions, even with incomplete modality information.

\end{itemize}

\section{Preliminaries}

\begin{figure}
\begin{center}
   \includegraphics[width=1\linewidth]{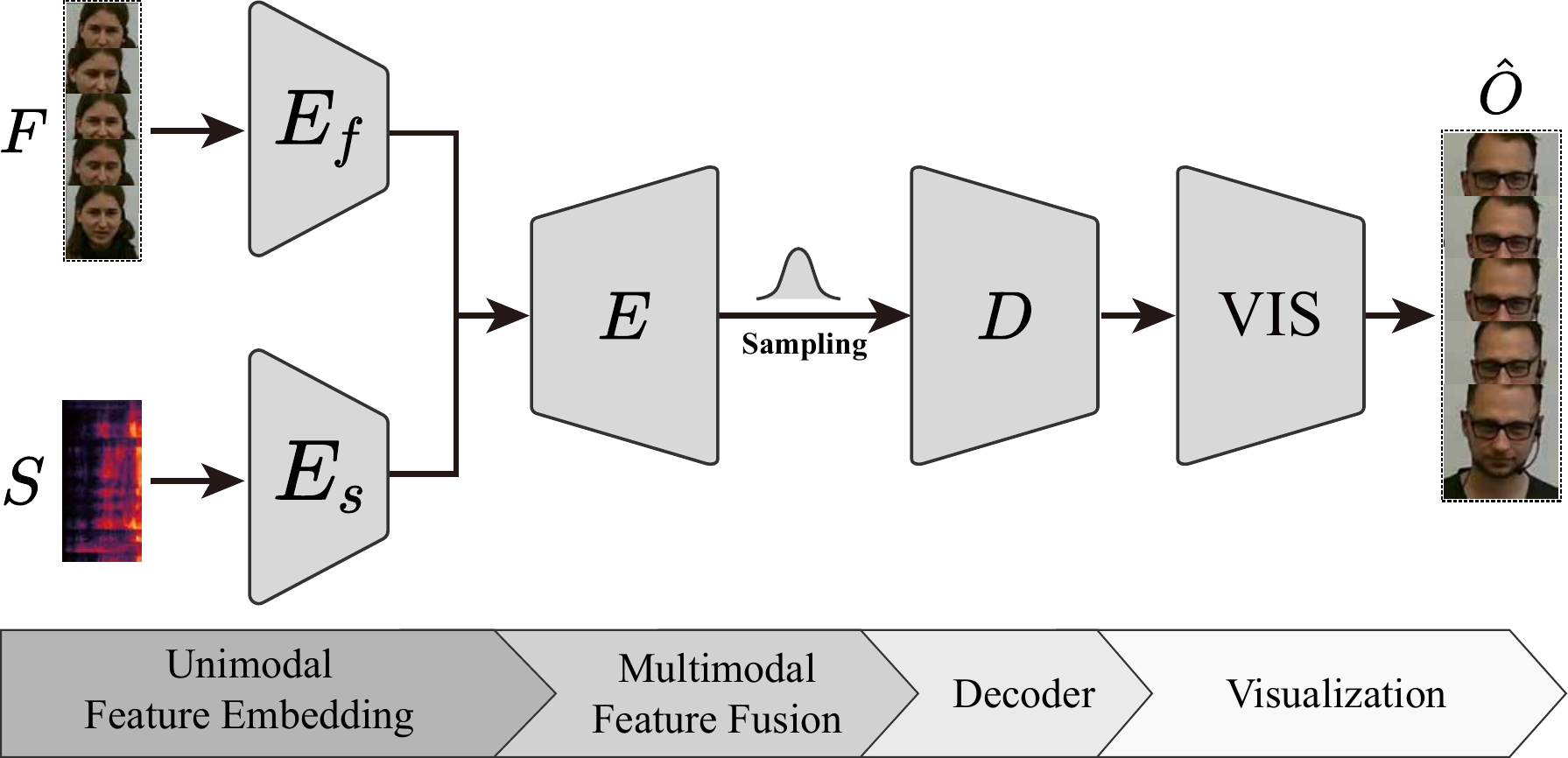}
\end{center}
   \caption{Example of a standard MAFRG model, which comprises four stages:  Unimodal Feature Encoding Stage, Multimodal Feature Fusion Stage, Decoding Stage, and Visualisation Stage.}
\label{fig:1}
\end{figure}

Most existing MAFRG models typically encompass four stages, as illustrated in Figure~\ref{fig:1}: Unimodal Feature Encoding Stage, Multimodal Feature Fusion Stage, Decoding Stage, and Visualisation Stage. This structure ensures the learning of appropriate and diverse listener facial representations \cite{react2023MM}. The Unimodal Feature Encoding Stage generally corresponds to the visual/audio unimodal encoder $E_f/E_s$, the Multimodal Feature Fusion Stage to the multimodal fusion module $E$, the Decoding Stage to the multiple appropriate facial reaction representations decoder $D$, and finally, the Visualisation Stage to a visualisation module $\textit{VIS}$. This section delineates this standard model structure.

To process multi-modal inputs, such as facial and speech data, the multimodal feature encoding stage of existing models generally accepts two primary inputs: the facial sequences $ F = \{f_1, f_2, \ldots, f_T\} $ and the speech sequences $ S = \{s_1, s_2, \ldots, s_T\} $, where $ f_t \in \mathbb{R}^{d_f} $ and $ s_t \in \mathbb{R}^{d_s} $, with $ d_f $ and $ d_s $ denoting the dimensions of the facial and speech inputs, respectively. Specifically, \( d_f = H \times W \times C \) and \( d_s = L \times M \), where \( H \), \( W \), and \( C \) represent the height, width, and number of channels of the facial input, respectively, while \( L \) and \( M \) denote the length and feature dimensions of the speech input. The facial sequence $ F $ can represent the facial state of the speaker in various formats, such as facial frame images or 3D Morphable Face Models (3DMM) vectors. Similarly, the speech sequence \( S \) consists of clip-level speech descriptors such as Mel-Frequency Cepstral Coefficients (MFCCs) or features derived from speech embedding models (e.g., Wav2Vec). To encode these input facial and speech sequences into their corresponding unimodal embeddings \( u^f_t \) and \( u^s_t \), there are dedicated facial and speech encoders, denoted as \( E_f \) and \( E_s \), respectively:

\begin{equation}
 \begin{split}
    u^f_t = E_f(f_t) , \quad t = 1, 2, \ldots, T \\
    u^s_t = E_s(s_t) , \quad t = 1, 2, \ldots, T   
\end{split}   
\end{equation}
where $ u^f_t, u^s_t \in \mathbb{R}^{d_u} $, and $ d_u $ represents the dimension of the embeddings.

Susequently, In the multimodal feature fusion stage, these encoded features are fed into a Multimodal Fusion Module $E$, which typically consists of two sequential parts. The first part is the multimodal feature fusion component $ E_m $, which fuses the facial and speech features to produce a multimodal speaker feature vector:

\begin{equation}
m_t = E_m(u^f_t, u^s_t) , \quad t = 1, 2, \ldots, T
\end{equation}

where $ m_t \in \mathbb{R}^{d_m} $ and $ d_m $ represents the dimension of the fused feature vector. The second part of the Multimodal Fusion Module is the distribution learning process, which learns the distribution $\mathbb{Z}$ of multiple suitable facial reactions corresponding to the input speaker behaviour. This process can be represented as:
\begin{equation}
    \mathbb{Z} = E_p(m_1, m_2, \ldots, m_T)
\end{equation}
where $ E_p $ denotes the prediction function that maps the fused multimodal feature vectors to the distribution of appropriate facial reactions.

Then, the decoder \( D \) in the decoding stage processes a set of samples \( g = \{g_1, g_2, \ldots, g_T\} \), where \( g_t \sim \mathbb{Z} \), to describe appropriate facial reactions according to the predicted distribution \(\mathbb{Z}\):

\begin{equation}
\hat{l}_t = D(g_t) , \quad t = 1, 2, \ldots, T
\end{equation}
where $l_t$ represents the decoded features corresponding to the listener's appropriate facial reactions.

Finally, in the visualisation stage, these predicted features are subsequently used to generate the listener's reaction frame sequences $ \hat{O} = \{\hat{o}_1, \hat{o}_2, \ldots, \hat{o}_T\} $ through the visualisation model $ Vis $:
\begin{equation}
\hat{o}_t = Vis(\hat{l}_t) , \quad t = 1, 2, \ldots, T
\end{equation}

\section{Methodology}

\subsection{Emotion-aware Modality Compensatory (EMC) Framework}

\begin{figure*}
\begin{center}
   \includegraphics[width=0.9\linewidth]{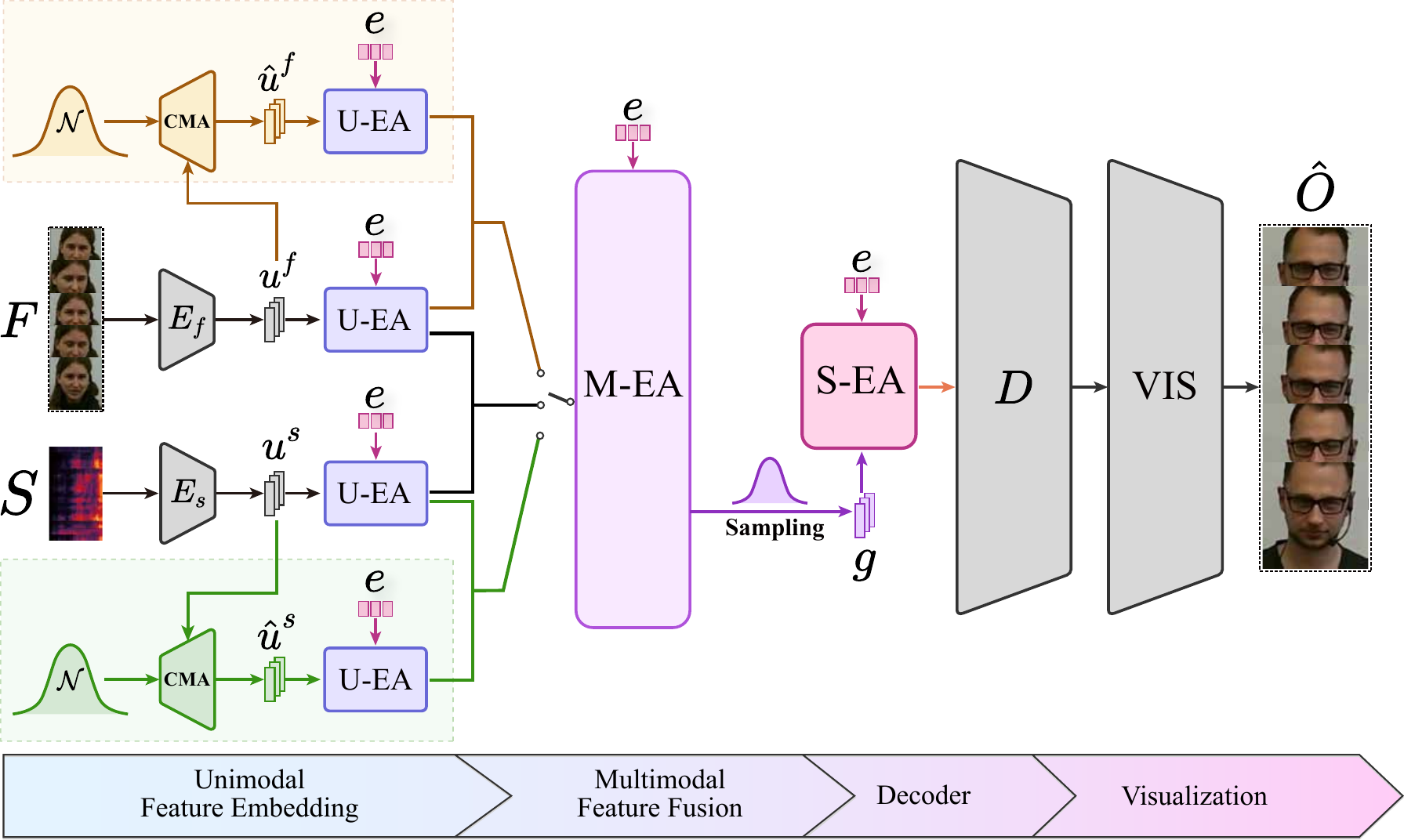}
\end{center}
   \caption{Architecture of the Emotion-aware Modality Compensatory (EMC) Framework.}
\label{fig:2}
\end{figure*}


We propose an Emotion-aware Modality Compensatory (EMC) framework, which comprises four stages, following the same workflow as the standard MAFRG model structure. Our framework primarily enhances the first three stages: the Unimodal Feature Encoding Stage, the Multimodal Feature Fusion Stage and the Decoding Stage. The complete architecture of the EMC is depicted in Figure \ref{fig:2}.

\vspace{-0.4cm} 
\subsubsection{Unimodal Feature Encoding Stage}




In the Unimodal Feature Encoding Stage, we introduced the Compensatory Modality Alignment (CMA) module to address the issue of modality absence and the Unimodal Emotion-aware Attention (U-EA) module to insert emotional cues into the unimodal feature embeddings.

The CMA Module maps a distribution to represent the speech or facial latent space, the details of the CMA module are introduced in section 4.2. When one data modality is missing, the corresponding CMA module compensates by providing a reliable substitute from the distribution. This ensures robust and coherent feature representations. We implement two CMA modules to learn a CMA Compensatory distribution $\mathcal{N}(\mu, \sigma^2)$ that encapsulates the speaker's unimodal information (speech or facial). During the inference phase, in scenarios where modality data is absent, we sample from this learned distribution as a substitute for the missing modality features. Let $ z \sim \mathcal{N}(\mu, \sigma^2)  $ be a sample from the distribution. For the case where the speech modality is missing, the CMA generates the missing speech substitute embedding $ \hat{u}^s_t $ as follows:

\begin{equation}
\hat{u}^s_t = \textit{CMA}(z^s_t, u^f_t), \quad t = 1, 2, \ldots, T \end{equation}

Similarly, for the case where the facial modality is missing, the CMA generates the missing facial substitute embedding $ \hat{u}^f_t $ as follows:

\begin{equation}
    \hat{u}^f_t = \textit{CMA}(z^f_t, u^s_t), \quad t = 1, 2, \ldots, T  
\end{equation}
where $ u^f_t $ and $ u^s_t $ are the embedding of the facial and speech features, respectively. 


Following the unimodal feature encoding, we introduce the Unimodal Emotion-aware Attention (U-EA) module. This module applies our proposed Emotion-aware Attention (EA) mechanism to all uni-modality features (i.e. speech and facial and CMA features) and ensures that the emotional nuances in the input data are more precisely captured, comprehended, and represented in the subsequent multimodal feature fusion process, the detail of EA are introduced in section 4.3. Mathematically, for the facial modality, the emotionally enhanced unimodal embedding $e^f_t$  is obtained by:

\begin{equation}
e^f_t = \textit{U-EA}(u^f_t, e_t), \quad t = 1, 2, \ldots, T 
\end{equation}
where $ e_t $ is the emotion feature at time step $t$ extracted from $f_t$. Similarly, for the speech modality, the enhanced embedding $ e^s_t $ is obtained by:

\begin{equation}
e^s_t = \textit{U-EA}(u^s_t, e_t), \quad t = 1, 2, \ldots, T
\end{equation}

In scenarios where the facial/speech modality data is missing, we use the Compensatory facial/speech feature embedding $ \hat{u}^f_t $/$ \hat{u}^s_t $  generated by the CMA module. The emotional enhanced CMA embedding $ \hat{e}^f_t $/ $\hat{e}^s_t$ for the missing modality is obtained by:

\begin{equation}
\begin{split}
\hat{e}^f_t = \textit{U-EA}(\hat{u}^f_t, e_t), \quad t = 1, 2, \ldots, T \\ 
\hat{e}^s_t = \textit{U-EA}(\hat{u}^s_t, e_t), \quad t = 1, 2, \ldots, T 
\end{split}
\end{equation}
where $ e_t $ represents the emotion features extracted from the existing modality contexts. 

\vspace{-0.3cm} 
\subsubsection{Multimodal Feature Fusion Stage}
In the multimodal feature fusion stage, the Multimodal Fusion Emotion-aware Attention (M-EA) module is introduced to integrate emotional information into the fused features. The structure of M-EA is the same as the U-EA module, with the key difference being that M-EA can accept three types of fused features: speech + facial, $\mathcal{N}$ + speech, and $\mathcal{N}$ + facial. During training, these three types of features are randomly provided to the M-EA module to enhance its generalizability and robustness. Mathematically, the distributions learned from the M-EA module can be represented as follows:
\begin{itemize}
    \item For the facial and speech features:
    \begin{equation}
    \mathbb{Z}_{sf} = \textit{M-EA}(e^{f}_t, e^{s}_t), \quad t = 1, 2, \ldots, T 
    \end{equation}

    \item For the CAM facial features and speech features:
    \begin{equation}
    \mathbb{Z}_{\hat{s}f} = \textit{M-EA}(\hat{e}^f_t, e^{s}_t), \quad t = 1, 2, \ldots, T 
    \end{equation}

    \item For the facial features and CAM speech features:
    \begin{equation}
        \mathbb{Z}_{s\hat{f}} = \textit{M-EA}(e^{f}_t, \hat{e}^s_t), \quad t = 1, 2, \ldots, T 
    \end{equation}
    where $\mathbb{Z}$ denotes the predicted listener's facial reaction distributions. 
\end{itemize}

To ensure that these three distributions remain consistent during the training process, we use a Kullback-Leibler (KL) divergence loss function that measures the consistency between these distributions. Let $ D_{\text{KL}} $ denote the KL divergence between distributions. We can define a fusion loss $ \mathcal{L}_{fusion} $ to minimize the differences between the three distributions. The loss function can be expressed as:

\vspace{-0.5cm} 

\begin{multline}
\mathcal{L}_{fusion} = \sum_{t=1}^{T} \Big( D_{\text{KL}}(\mathbb{Z}_{sf} \parallel \mathbb{Z}_{\hat{s}f}) \Bigg. \\
\Bigg. + D_{\text{KL}}(\mathbb{Z}_{sf} \parallel \mathbb{Z}_{s\hat{f}}) + D_{\text{KL}}(\mathbb{Z}_{\hat{s}f} \parallel \mathbb{Z}_{s\hat{f}}) \Big)
\end{multline}


\vspace{-0.5cm} 
\subsubsection{Decoding and Visualisation Stage}

In the decoding stage, the Sample Emotion-aware (S-EA) Attention module plays a crucial role in refining the generated samples to ensure that the generated listener responses accurately reflect the emotional tone and context of the speaker. 
Let $ g_t \sim \mathbb{Z} $ represent the sample generated from the distribution $ \mathbb{Z} $ at time step $ t $. The S-EA module integrates the emotional context $ e_t $ into the samples to produce refined samples $ g^{emo}_t $ as:
\begin{equation}
g^{emo}_t = \textit{S-EA}(g_t, e_t), \quad t = 1, 2, \ldots, T 
\end{equation}
where $e_t$ represents the emotional features extracted from the context. 

Finally, the emotionally refined samples can be decoded using the standard model's Decoder and Visualiser to generate a more appropriate response. 

\subsection{Compensatory Modality Alignment Module}

\begin{figure}
\begin{center}
   \includegraphics[width=1\linewidth]{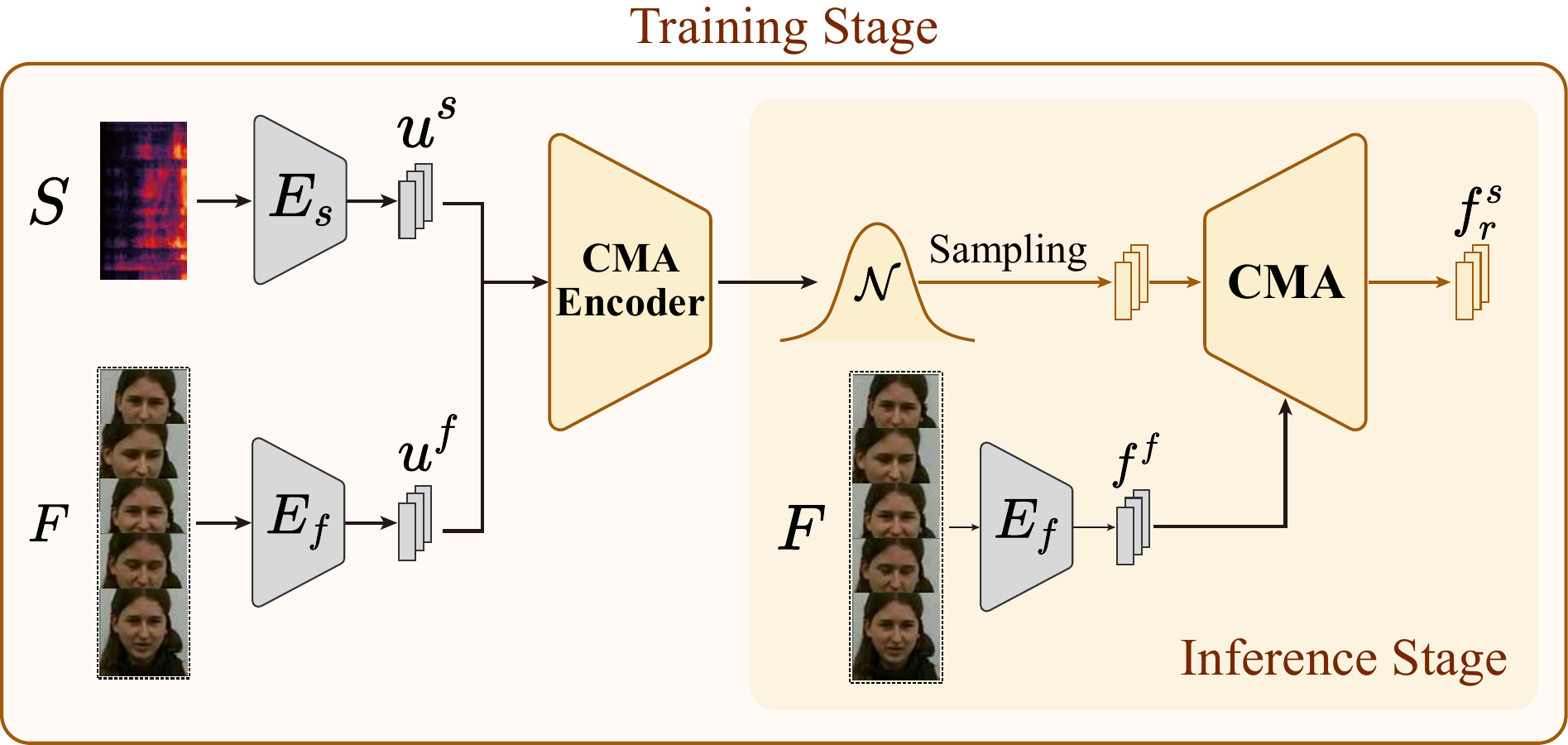}
\end{center}
   \caption{The structure of the Compensatory Modality Alignment (CMA) module in the training and inference stages for the scenario where speech modality data is missing.}
\label{fig:4}
\end{figure}

To address the challenge of performance degradation in the standard MAFRG model due to missing modalities, we introduce the Compensatory Modality Alignment (CMA) Module, inspired by Audio2Gestures \cite{li2021audio2gestures}. The CMA module is designed to generate representations of missing modalities from the learned Compensatory distribution, ensuring effective handling of incomplete multimodal data during inference. In our framework, two CMA modules are integrated within the Unimodal Feature Encoding stage—the Facial CMA and the Speech CMA—each tasked with generating corresponding modality representation substitutes. Using the Speech CMA as an example, the training and inference stages are illustrated in Figure~\ref{fig:4}.

During the inference phase, when speech modality data is absent, we can sample from the speech CMA distribution and use the CMA module to generate Compensatory features. These samples are conditioned on the available facial features to ensure precise alignment. This mechanism maintains robust performance and preserves the integrity of multimodal representations within the latent space.

To learn the CMA Compensatory distribution, we utilise a Conditional Variational Autoencoder (CVAE) structure, which consists of an encoder and a decoder. The encoder (CMA Encoder) receives both facial feature \( u^f \) and speech features \( u^s \) as input. The speech features are used to learn their distribution, while the facial features \( u^f \) are used as a condition. The encoder outputs the mean \( \mu_s \) and standard deviation \( \sigma_s \) of the distribution \( z_s \), which characterizes the distribution \( \mathcal{N}_s(\mu_s, \sigma_s) \). These samples are then input into the decoder CMA module to generate the modality-compensatory feature \(\hat{u}^s\).

The training objective comprises two main terms: the latent space alignment loss and the KL divergence loss. The latent space alignment loss ensures that the generated feature \(\hat{u}^s\) aligns with the features obtained from the speech encoder \( u^s \):
\vspace{-0.5cm} 

\begin{equation}
\mathcal{L}_{align} = \sum_{i=1}^{n} \| u^s_i - \hat{u}^s_i \|^2
\end{equation}
where \( n \) is the number of samples. The KL divergence regularises the learned distribution to be close to the Gaussian distribution:
\begin{equation}
\mathcal{L}_{KL} = D_{KL}(\mathcal{N}_s(\mu_s, \sigma_s) \| \mathcal{N}(0, I))
\end{equation}
Thus, the overall loss function for training the CVAE combines these two terms:
\begin{equation}
\mathcal{L}_{CVAE} = \mathcal{L}_{align} + \mathcal{L}_{KL}
\end{equation}

\subsection{Emotion-aware Attention}

\begin{figure}
\begin{center}
   \includegraphics[width=1\linewidth]{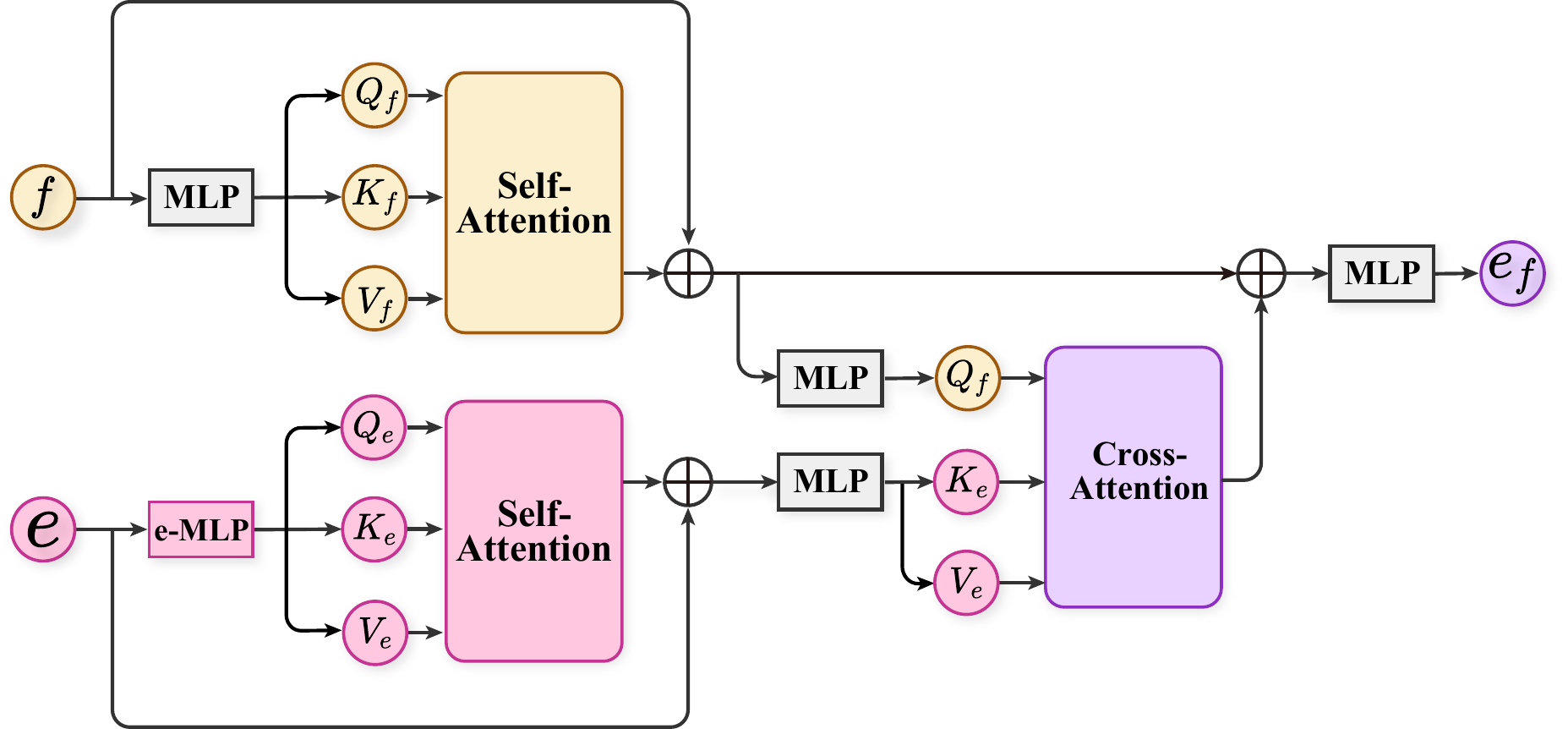}
\end{center}
   \caption{The structure of Emotion-aware Attention (EA) module}
\label{fig:3}
\end{figure}





To integrate the speaker's emotional context into the listener's response generation, we present the Emotion-aware Attention (EA) Module, which modulates attention based on emotional context. This enhancement facilitates emotional understanding and generates contextually appropriate responses. The architecture of the EA module is illustrated in Figure~\ref{fig:3}.

The EA module comprises two self-attention components and a cross-attention component. Initially, modality features $f$ and emotion embeddings $e$ are processed separately through their respective self-attention blocks to capture temporal dependencies and enhance feature representation. Subsequently, the refined modality features serve as the Query ($Q_f$), while emotion embeddings provide the Key ($K_e$) and Value ($V_e$) inputs to the cross-attention block. This cross-attention mechanism dynamically aligns input features with emotion vectors, integrating external emotional information and directing the model's attention towards the most relevant emotional aspects. By refining attention from both self-attention and cross-attention layers with emotion-specific adjustments, the EA module ensures emotionally nuanced outputs.

The proposed EA module is a versatile component capable of receiving various inputs, including unimodal features, CMA features, and speech-facial fusion features. It can be seamlessly integrated into different models to enhance emotional awareness. Notably, the emotion self-attention component can accept emotion features extracted from either facial or speech data, thanks to our meticulously designed e-MLP module. In cases where only the facial modality is available, we follow the REACT methodology \cite{react2023arxiv}, extracting facial features from the state-of-the-art facial emotion recognition model. This includes the occurrence of 15 facial action units (AUs), valence and arousal intensities, and the probabilities of 8 categorical facial expressions. Conversely, when only the speech modality is present, we extract emotion features using a speech emotion model, focusing on valence and arousal intensities, and the probabilities of 8 categorical speech expressions. This approach ensures high adaptability to scenarios where certain modalities may be missing.

This flexible design ensures that the EA module effectively integrates and prioritizes emotional cues from diverse sources, enabling the model to respond appropriately even when certain input modalities are absent. By emphasizing emotionally charged information during the encoding and decoding phases, the EA module guarantees that the generated responses are both contextually appropriate and emotionally resonant. This adaptability is particularly valuable in applications requiring emotional intelligence.

\section{Experiments}


\subsection{Experiment settings}



Following the REACT 2024 offline challenge rules \cite{song2024react}, we focus on predicting facial reactions to a pre-recorded segment of the speaker’s behaviour. To evaluate our proposed strategy, we utilise three state-of-the-art (SOTA) representative MAFRG models as baseline models: Trans-VAE \cite{song2024react}, REGNN \cite{xu2023reversible}, and FaceAIS \cite{damfinite}. These models are employed to integrate our EMC. All models were implemented using the PyTorch framework and executed on servers equipped with one NVIDIA Tesla V100 GPU. For training parameters, including the optimizer and learning rate, we adhered to the experimental protocols used in the original baseline models. This ensured a fair comparison and validated the performance improvements and efficiency in handling missing modality data in facial reaction generation brought by our EMC framework. We evaluated the performance of the proposed CMA
framework on the REACT2024 corpus \cite{song2024react}, more details are provided in the Supplementary Material.

\subsection{Evaluation metrics}

We evaluated the performance based on the six metrics defined by REACT 2024 \cite{song2024react,noxi,recola}, which assess the comprehensive performance of facial reaction generation across three aspects: Appropriateness, Diversity, and Synchrony. Two metrics for evaluating appropriateness are FRDist and FRCorr. FRDist measures the distance between the generated facial expressions and the ground-truth facial expressions, while FRCorr assesses the correlation between them. The diversity of generated facial reactions is evaluated using FRVar, FRDiv, and FRDvs, which compute the variation among the reactions. Additionally, Time Lagged Cross Correlation (referred to as FRSyn) is employed to assess synchrony. More detailed formulations and explanations are in Supplementary.
\subsection{Experimental results}

\begin{table*}[t]
\begin{center}
\renewcommand{\arraystretch}{1.2} 
\resizebox{\textwidth}{!}{
\begin{tabular}{|c|c|cc|ccc|c|}
\hline
                                           &                                   & \multicolumn{2}{c|}{\textbf{Appropriateness}}              & \multicolumn{3}{c|}{\textbf{Diversity}}                                                       & \textbf{Synchrony}           \\ 
\multirow{-2}{*}{\textbf{Models}}          & \multirow{-2}{*}{\textbf{Method}} & \textbf{FRCorr ↑}           & \textbf{FRDist ↓}            & \textbf{FRDiv ↑}              & \textbf{FRVar ↑}              & \textbf{FRDvs ↑}              & \textbf{FRSyn ↓}             \\ \hline
                                           & \textbf{Trans-VAE \cite{song2024react}}                & 0.03                        & 92.81                        & 0.0008                        & 0.0002                        & 0.0006                        & 43.75                        \\ 
                                           & \textbf{REGNN \cite{xu2023reversible}}                    & 0.19                        & 84.54                        & 0.0007                        & 0.0061                        & 0.0342                        & 41.35                        \\ 
\multirow{-3}{*}{\textbf{Original}}        & \textbf{FaceVIS(*) \cite{damfinite}}               & 0.23                        & 88.43                        & 0.1158                        & 0.0346                        & 0.1159                        & 44.99                        \\ \hline
                                           & \textbf{Trans-VAE-EMC}                & {\color[HTML]{3166FF} 0.07} & {\color[HTML]{3166FF} 31.30} & {\color[HTML]{3166FF} 0.0454} & {\color[HTML]{3166FF} 0.0201} & {\color[HTML]{3166FF} 0.0495} & {\color[HTML]{3166FF} 42.63} \\ 
                                           & \textbf{REGNN-EMC}                    & {\color[HTML]{3166FF} 0.23} & 93.33                        & 0.0007                        & 0.0056                        & {\color[HTML]{3166FF} 0.0445} & {\color[HTML]{3166FF} 39.02} \\ 
\multirow{-3}{*}{\textbf{Ours (EMC)}}      & \textbf{FaceVIS-EMC}                  & {\color[HTML]{3166FF} 0.27} & {\color[HTML]{3166FF} 86.37} & 0.1078                        & {\color[HTML]{3166FF} 0.0619} & 0.1077                        & {\color[HTML]{3166FF} 43.82} \\ \hline
                                           & \textbf{Trans-VAE-EMC}                & 0.03                        & {\color[HTML]{3166FF} 32.31} & {\color[HTML]{3166FF} 0.0378} & {\color[HTML]{3166FF} 0.0209} & {\color[HTML]{3166FF} 0.042}  & 45.61                        \\ 
                                           & \textbf{REGNN-EMC}                    & {\color[HTML]{3166FF} 0.21} & 90.75                        & 0.0006                        & 0.0054                        & {\color[HTML]{3166FF} 0.0405} & {\color[HTML]{3166FF} 39.92} \\ 
\multirow{-3}{*}{\textbf{Ours w/o EA}}     & \textbf{FaceVIS-EMC}                  & 0.23                        & 90.32                        & 0.1088                        & {\color[HTML]{3166FF} 0.0495} & 0.1089                        & {\color[HTML]{3166FF} 44.89} \\ \hline
                                           & \textbf{Trans-VAE-EMC}                & 0.02                        & {\color[HTML]{3166FF} 42.89} & {\color[HTML]{3166FF} 0.0022} & {\color[HTML]{3166FF} 0.0005} & {\color[HTML]{3166FF} 0.0072} & 45.79                        \\ 
                                           & \textbf{REGNN-EMC}                    & 0.11                        & {\color[HTML]{3166FF} 83.55} & 0.0006                        & 0.0039                        & 0.0127                        & {\color[HTML]{3166FF} 39.75} \\ 
\multirow{-3}{*}{\textbf{Ours w/o Facial}} & \textbf{FaceVIS-EMC}                  & 0.19                        & 97.60                        & 0.0839                        & 0.0295                        & 0.0839                        & {\color[HTML]{3166FF} 44.6}  \\ \hline
                                           & \textbf{Trans-VAE-EMC}                & {\color[HTML]{3166FF} 0.05} & {\color[HTML]{3166FF} 31.83} & {\color[HTML]{3166FF} 0.0209} & {\color[HTML]{3166FF} 0.0179} & {\color[HTML]{3166FF} 0.0133} & 46.66                        \\ 
                                           & \textbf{REGNN-EMC}                    & {\color[HTML]{3166FF} 0.22} & 92.65                        & 0.0006                        & 0.0051                        & {\color[HTML]{3166FF} 0.0408} & {\color[HTML]{3166FF} 40.68} \\ 
\multirow{-3}{*}{\textbf{Ours w/o Speech}} & \textbf{FaceVIS-EMC}                  & 0.23                        & {\color[HTML]{3166FF} 87.85} & 0.0856                        & 0.0235                        & 0.0857                        & 45.25                        \\ \hline
\end{tabular}
}
\end{center}
\caption{Comparison of Models on Appropriateness, Diversity, and Synchrony Metrics (Blue indicates improvement over the original results; * indicates results obtained from official pre-trained models)}
\label{tab:1}
\end{table*}




Table \ref{tab:1} presents a comparative analysis of the effectiveness of our proposed EMC framework. \textbf{Original} denote the three existing SOTA models we selected, while \textbf{Ours} are the corresponding optimised versions based on our proposed EMC framework. \textbf{Ours w/o EA} represents the EMC framework without the Emotion Attention module,\textbf{Ours w/o Facial} shows the performance of our EMC when facial data is absent. and \textbf{Ours w/o Speech} shows the performance of our EMC when speech data is absent. 

\vspace{-0.3cm} 
\subsubsection{Superiority of the Proposed EMC Framework}
Our EMC framework significantly outperforms the original models across multiple metrics, demonstrating its effectiveness in generating appropriate, diverse, and synchronised facial reactions. Notably, the Trans-VAE method with EMC showed substantial improvements: FRCorr increased by 133.3\%, reflecting a stronger correlation between generated and ground-truth facial expressions, while FRDist decreased by 66.3\%, indicating a significant reduction in discrepancy. The diversity metrics also saw remarkable gains, with FRDiv increasing by 5575\%, FRVar by 9950\%, and FRDvs by 8150\%, highlighting the model's enhanced capability to produce varied facial reactions.

In the case of FaceVIS, when enhanced with EMC, there is a 17.4\% improvement in FRCorr and a only 2.3\% decrease in FRDist, affirming the framework's ability to generate more appropriate facial reactions. Despite a slight 6.9\% decrease in FRDiv, the FRVar increased by 78.9\%, indicating more nuanced variations among the reactions. The REGNN model also demonstrated significant enhancements, with a 21.1\% improvement in FRCorr and a 30.1\% increase in FRDvs, although FRDist increased by 10.4\% and FRVar experienced a minor reduction.

The synchrony metric (FRSyn) showed minor changes across all models, with slight decreases that did not significantly impact the alignment of reactions with speaker's behaviour. This cohesive analysis underscores the robustness and enhanced performance of our EMC framework in facial reaction generation tasks, proving its superiority in improving appropriateness, diversity, and synchrony.



\vspace{-0.2cm} 
\subsubsection{Model Performance with Missing Modality Data}

Unlike the original models, which tend to fail in scenarios where modality data are missing, our proposed framework is capable of generating facial reactions even under such conditions, highlighting the resilience and adaptability of our approach.

In scenarios where speech data are missing, our framework demonstrates notable performance advantages in specific metrics. For instance, Trans-VAE-EMC, without speech data, achieved an FRCorr of 0.05, representing a 66.67\% improvement over the original Trans-VAE's 0.03, which utilised both speech and facial data. Similarly, REGNN-EMC, in the absence of speech data, attained an FRCorr of 0.22, marking a 15.8\% increase compared to the original REGNN's 0.19. These enhancements underscore the capability of our CMA to compensate for the missing modality by effectively leveraging the available modalities.

Without facial data, the Trans-VAE-EMC model's FRCorr dropped to 0.02, a 33.3\% decrease from the original model. The FaceVIS model exhibited a similar trend, where the absence of facial data led to a decline in performance metrics such as FRDist and FRDvs. Notably, the REGNN-EMC, even without facial data, managed to improve its FRDist by 1.2\%, demonstrating that while facial data are crucial, the model can still compensate for their absence using our EMC framework. This highlights the effectiveness of our CMA module, which aligns and integrates available data to mitigate the impact of missing information.

The observed performance differences when either speech or facial data are missing stem from the unique information each modality provides. Facial data directly conveys visual emotional expressions, essential for accurate facial reactions, leading to a significant drop in performance metrics like FRCorr and FRDist when missing. In contrast, speech data provides complementary information about tone, pitch, and emotional state, impacting performance less drastically. The Compensatory Modality Alignment (CMA) module plays a crucial role in maintaining performance when data is missing. It effectively infers and reconstructs missing data by leveraging the available modality. This is particularly effective with speech data, as its less complex and more uniform nature allows for easier compensation and integration. Consequently, the CMA module enhances the robustness and effectiveness of the framework by accurately reconstructing missing speech data.

\begin{figure*}
\begin{center}
   \includegraphics[width=1\linewidth]{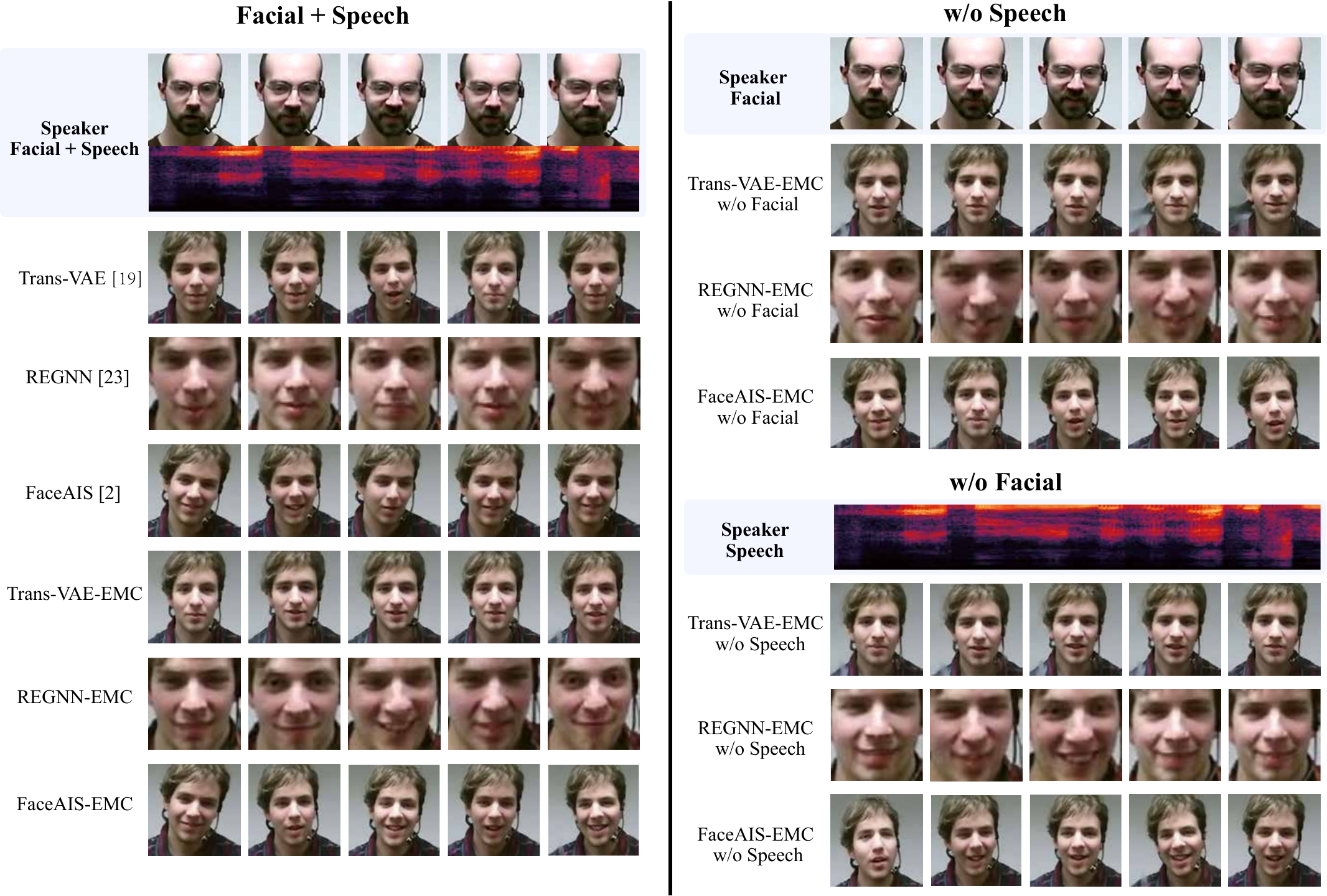}
\end{center}
   \caption{Visualisation of the facial reactions generated from different approaches. It shows that our EMC-enhanced methods can generate more appropriate facial reactions; In scenarios where speech modality data is missing (w/o Speech), the performance of appropriate generation shows an improvement, and when facial data is missing, it only exhibits minimal degradation  (w/o Facial).}
\label{fig:5}
\end{figure*}

The visualizations in Figure \ref{fig:5} showcase the detailed facial reactions generated by our framework. Our enhanced models, such as Trans-VAE-EMC and REGNN-EMC, produce nuanced and accurate facial expressions that closely mirror the emotional tone and context of the speaker's inputs. For example, in the absence of facial data, Trans-VAE-EMC can still generate subtle eyebrow movements and mouth shapes that reflect the speaker's emotional state, thanks to the effective reconstruction capabilities of the CMA module. Similarly, REGNN-EMC, despite missing facial input, maintains consistent eye and lip emotion synchronization, ensuring that the generated reactions appear natural and coherent.



\vspace{-0.2cm} 

\subsection{Ablation Study}
\vspace{-0.1cm} 
To further understand the impact of our proposed modules, we performed ablation studies by removing the EA Module. This analysis aimed to evaluate how critical the incorporation of emotional cues is to the overall performance of our EMC framework. As expected, the removal of the EA module led to noticeable decreases in the appropriateness of the generated facial reactions. Specifically, the FRCorr score of the Trans-VAE-EMC model dropped by 133\% compared to the version with the EA module, highlighting the significant role that emotional information plays in enhancing the correlation between generated and ground-truth facial expressions. 

Furthermore, the diversity metrics—FRDiv, FRVar, and FRDvs—showed no significant improvement or deterioration compared to the complete EMC framework. This indicates that while the EA module profoundly impacts the appropriateness of facial reactions, its influence on the diversity of generated reactions is minimal. The generated reactions' variation and richness remained consistent whether the EA module was included or not. This observation leads us to hypothesise that additional emotional information provided by the EA module primarily enhances the contextual relevance and appropriateness of reactions rather than contributing to the diversity of reaction generation. The diversity of reactions appears to be more influenced by the compensatory mechanisms for handling missing modality data.

\section{Conclusion \& Future Work}

In this paper, we introduce the EMC framework to address the limitations of existing models that do not adequately consider the importance of speaker emotion cues in facial reaction generation. Additionally, these models generally struggle with scenarios involving missing modality data. Our framework introduces the CMA and EA modules, ensuring robust and contextually appropriate facial reactions even when some data modalities are unavailable. Experimental results demonstrate significant improvements in both the appropriateness and diversity of the generated reactions. The EMC framework’s resilience and enhanced performance make it a valuable contribution to advancing human-computer interaction, ensuring more natural and emotionally attuned communication between virtual agents and humans.

As future work, we intend to integrate more emotional and affective states in our module \cite{kollias2023facernet,kollias2023multi,kollias2024distribution,psaroudakis2022mixaugment,kollias2021distribution,kollias2019face}. 


{\small
\bibliographystyle{ieee}
\bibliography{egbib}
}

\end{document}


\title{Supplementary }

\maketitle
\thispagestyle{empty}

\section{Datasets}

We evaluated the performance of the proposed CMA framework on the REACT2024 corpus \cite{song2024react}. The REACT2024 corpus comprises dyadic interaction video clips, incorporating data from two datasets: NoXi \cite{noxi} and RECOLA \cite{recola}. In total, REACT2024 consists of 5,924 video clips, each 30 seconds in length. We adhered to the official subject-independent evaluation strategy, which divides REACT2024 into 3,188 clips for training, 1,124 for validation, and 1,612 for testing.

\section{Evaluation Metrics}

\noindent \subsection{Appropriateness Metrics:}

\noindent 1. Facial Reaction Distance (FRDist):
   \[
   FRDist = \frac{1}{n} \sum_{i=1}^n \min_{j} \text{DTW}(R_{i,j}, G_{i,j})
   \]
   where \( n \) is the number of reactions, \( R_{i,j} \) represents the generated reaction, and \( G_{i,j} \) represents the ground truth reaction.

\vspace{1em}
\noindent 2. Facial Reaction Correlation (FRCorr):
   \[
   FRCorr = \frac{1}{n} \sum_{i=1}^n \max_{j} \text{CCC}(R_{i,j}, G_{i,j})
   \]
   where \( \text{CCC} \) stands for Concordance Correlation Coefficient.

\vspace{1em}
\noindent 3. Prediction Accuracy (ACC):
   \[
   ACC = \frac{1}{n} \sum_{i=1}^n \mathbf{1}(\min_{j} \text{DTW}(R_{i,j}, G_{i,j}) < \tau)
   \]
   where \( \mathbf{1} \) is the indicator function and \( \tau \) is the threshold.

\noindent \subsection{Diversity Metrics:}

\noindent 1. Facial Reaction Variance (FRVar):
   \[
   FRVar = \frac{1}{n} \sum_{i=1}^n \text{Var}(R_i)
   \]
   where \( \text{Var} \) denotes variance.

\vspace{1em}
\noindent 2. Diverseness (FRDiv):
   \[
   FRDiv = \frac{1}{n} \sum_{i=1}^n \sum_{j \neq k} \text{MSE}(R_{i,j}, R_{i,k})
   \]
   where \( \text{MSE} \) represents Mean Squared Error.

\vspace{1em}
\noindent 3. Inter-condition Diversity (FRDvs):
   \[
   FRDvs = \frac{1}{n} \sum_{i \neq k} \text{MSE}(R_i, R_k)
   \]

\noindent \subsection{Realism Metric:}

\noindent Facial Reaction Realism (FRRea):
   \[
   FRRea = \text{FID}(R, G)
   \]
   where \( \text{FID} \) stands for Fréchet Inception Distance.

\noindent \subsection{Synchrony Metric:}

\noindent Synchrony (FRSyn):
   \[
   FRSyn = \frac{1}{n} \sum_{i=1}^n \max_{\tau} \text{TLCC}(S_i, R_i)
   \]
   where \( \text{TLCC} \) represents Time-Lagged Cross-Correlation.

These metrics evaluate the facial reaction generation models by assessing appropriateness, diversity, realism, and synchrony with the input behavior. For more details, refer to the full paper \cite{whatwhyhow}.

{\small
\bibliographystyle{ieee}
\bibliography{egbib}
}